\documentclass[a4paper,twoside]{article}

\usepackage{epsfig}
\usepackage{graphicx}
\usepackage{subcaption}
\usepackage{calc}
\usepackage{amssymb}
\usepackage{amstext}
\usepackage{amsmath}
\usepackage{amsthm}
\usepackage{multicol}
\usepackage{placeins}
\usepackage{pslatex}
\usepackage{include/apalike}
\usepackage{paralist}
\usepackage{include/SCITEPRESS}     

\usepackage{paralist}
\usepackage[dvipsnames]{xcolor}
\usepackage{bm}
\usepackage{xspace}
\usepackage{booktabs}
\usepackage{multirow}
\usepackage{mathtools}

\usepackage{tikz}
\def\figspace{16pt}

\makeatletter

\DeclareRobustCommand\onedot{\futurelet\@let@token\@onedot}
\def\@onedot{\ifx\@let@token.\else.\null\fi\xspace}

\def\eg{\emph{e.g}\onedot} 
\def\ie{\emph{i.e}\onedot}

\DeclarePairedDelimiterX{\infdivx}[2]{(}{)}{%
  #1\;\delimsize\|\;#2%
}
\newcommand{\KL}{\text{KL}\infdivx}
\newcommand{\JSD}{\text{JSD}\infdivx}

\makeatother

\begin{document}

\title{Latent-Space Laplacian Pyramids for Adversarial \\
Representation Learning with 3D Point Clouds}


\author{\authorname{Vage Egiazarian\sup{* 1}, Savva Ignatyev\sup{* 1}, Alexey Artemov\sup{1},\\
Oleg Voynov\sup{1}, Andrey Kravchenko\sup{2}, Youyi Zheng\sup{3}, Luiz Velho\sup{4}, Evgeny Burnaev\sup{1}}
\affiliation{\sup{1}Skolkovo Institute of Science and Technology, Moscow, Russia}
\affiliation{\sup{2}DeepReason.ai, Oxford, UK}
\affiliation{\sup{3}State Key Lab, Zhejiang University, China}
\affiliation{\sup{4}IMPA, Brazil}
\email{\{vage.egiazarian, savva.ignatyev, a.artemov, o.voinov\}@skoltech.ru,\\
andrey.kravchenko@deepreason.ai, youyizheng@zju.edu.cn, lvelho@impa.br, e.burnaev@skoltech.ru}
}

\abstract{Constructing high-quality generative models for 3D shapes is a fundamental task in computer vision with diverse applications in geometry processing, engineering, and design. Despite the recent progress in deep generative modelling, synthesis of finely detailed 3D surfaces, such as high-resolution point clouds, from scratch has not been achieved with existing approaches. 
In this work, we propose to employ the latent-space Laplacian pyramid representation within a hierarchical generative model for 3D point clouds. We combine the recently proposed latent-space GAN and Laplacian GAN architectures to form a multi-scale model capable of generating 3D point clouds at increasing levels of detail. Our evaluation demonstrates that our model outperforms the existing generative models for 3D point clouds.}

\onecolumn \maketitle \normalsize \setcounter{footnote}{0} \vfill

\section{Introduction}
\label{sec:intro}

A point cloud is an ubiquitous data structure that has gained a strong presence in the last few decades with the widespread use of range sensors. Point clouds are sets of points in 3D space, commonly produced by range measurements with 3D scanners (\eg LIDARs, RGB-D cameras, and structured light scanners) or computed using stereo-matching algorithms. A key use-case with point clouds is 3D surface reconstruction involved in many applications such as reverse engineering, cultural heritage conservation, or digital urban planning.

Unfortunately, for most scanners the raw 3D measurements often cannot be used in their original form for surface/shape reconstruction, as they are generally prone to noise and outliers, non-uniform, and incomplete. Whilst constant progress in scanning technology has led to improvements in some aspects of data quality, others, such as occlusion, remain a persistent issue for objects with complex geometry. Thus, a crucial step in 3D geometry processing is to model full 3D shapes from their sampling as 3D point clouds, inferring their geometric characteristics from incomplete and noisy measurements. 
A recent trend in this direction is to apply data-driven methods such as deep generative models~\cite{achlioptas2018learning,li2018point,chen2019unpaired}.

\begin{figure}[t!]
  \centerline{
  \includegraphics[width=\columnwidth]
    {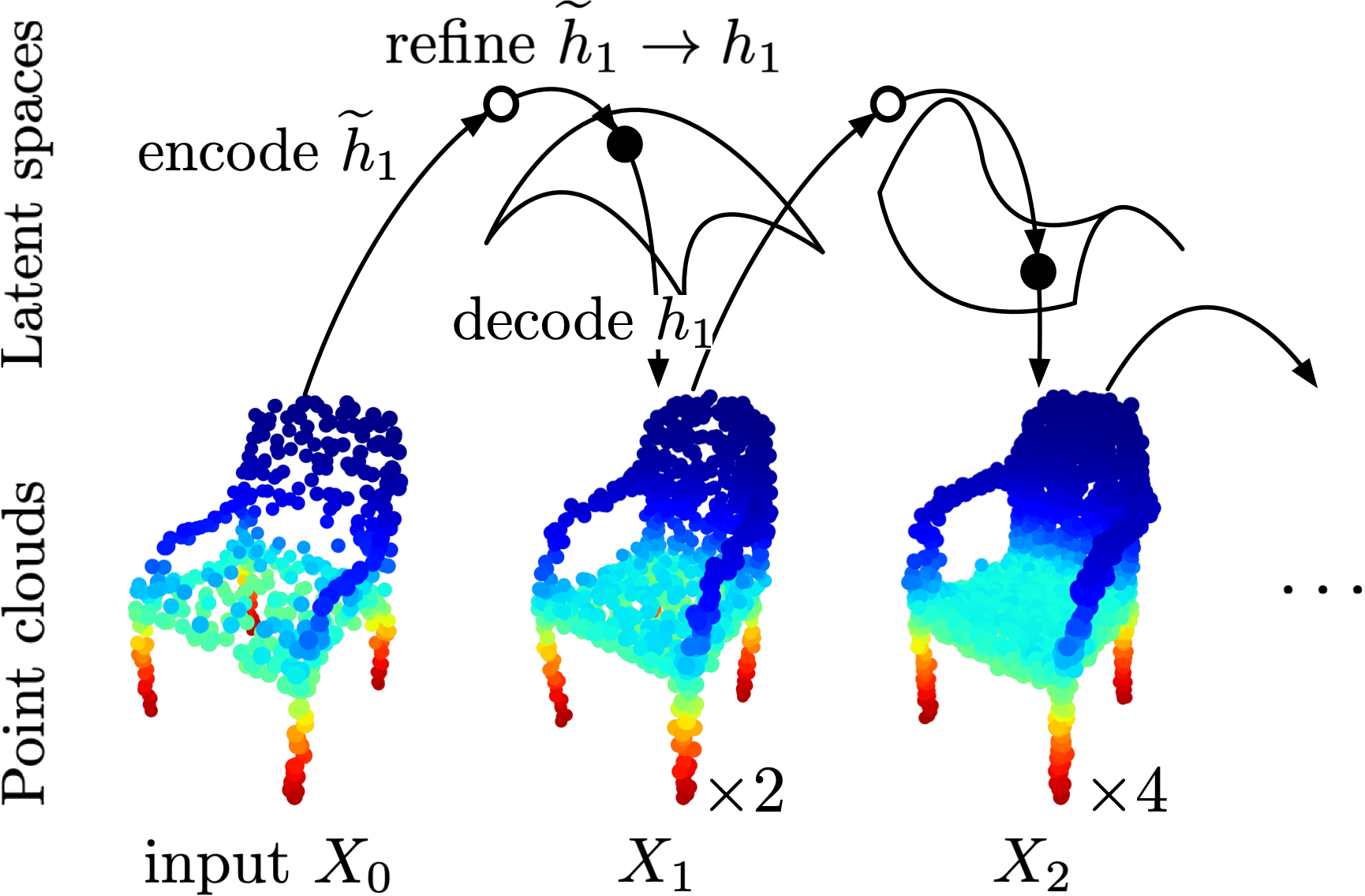}
  }
  \caption{The proposed latent-space Laplacian GAN operates on latent representations, bypassing the need to process large-scale 3D point clouds. Through using mutiple stages of the pyramid, the sample resolution can be iteratively increased.}
  \label{fig:teaser}
\end{figure}

However, most known deep models operate directly in the original space of raw measurements, which represents a challenging task due to the omnipresent redundancy in raw data; instead, it might be beneficial to encode characteristic shape features in the latent space and further operate with latent representations. Another shortcoming is that most models only operate with coarse (low-resolution) 3D geometry, as high-resolution 3D shapes are computationally demanding to process and challenging to learn from.

In this work, we consider the task of learning 3D shape representations given their 3D point cloud samples. To this end, we develop a novel deep cascade model, Latent-Space Laplacian Pyramid GAN, or LSLP-GAN, taking advantage of the recent progress in adversarial learning with point clouds~\cite{achlioptas2018learning} and deep multi-scale models~\cite{denton2015deep,mandikal2019dense}. Our proposed model (schematically represented in Figure~\ref{fig:teaser}) has several important distinctions: \begin{inparaenum}[(i)] \item it is generative and can be used to produce synthetic shapes unseen during training; \item it is able to produce high-resolution point clouds via a latent space Laplacian pyramid representation; \item it is easy to train as it operates in the space of latent codes, bypassing the need to perform implicit dimensionality reduction during training. We train our model in an adversarial regime using a collection of 3D point clouds with multiple resolutions. \end{inparaenum}


In this work, our contributions are the following.
\begin{enumerate}
    \item We propose LSLP-GAN, a novel multi-scale deep generative model for shape representation learning with 3D point clouds.
    
    \item We demonstrate by the means of numerical experiments the effectiveness of our proposed method for shape synthesis and upsampling tasks.
\end{enumerate}

\section{Related work}
\label{sec:related}

\noindent\textbf{Neural networks on unstructured 3D point clouds.}
\label{review:nn_point_clouds}
Despite Deep convolutional neural networks (CNNs) having  proved themselves to be very effective for learning with 2D grid-structured images, until very recently, the same could not be said about unsturctured 3D point clouds. The basic building blocks for point-based architectures (equivalents of the spatial convolution) are not straightforward to implement. To this end, MLP-based~\cite{Pointnet,Pointnet++}, graph convolutional~\cite{dgcnn}, and 3D convolutional point networks~\cite{pwcnn,pointcnn,pcnnExt} have been proposed, each implementing their own notion of convolution, and applied to classification, semantic labelling, and other tasks. We adopt the convolution of~\cite{Pointnet} as a basis for our architecture.

Building on top of the success of point convolutions, auto-encoding networks have been proposed~\cite{achlioptas2018learning,yang2018foldingnet,li2018so} to learn efficient latent representations. 

\noindent\textbf{Generative neural networks for 3D modelling. }
\label{review:generative_3d_data}
The literature on generative learning with 3D shapes shows instances of variational auto-encoders (VAEs)~\cite{kingma2013auto} and generative adversarial networks (GANs)~\cite{goodfellow2014generative} applied to 3D shape generation. On the one hand, VAEs have been demonstrated to efficiently model images~\cite{kingma2013auto}, voxels~\cite{brock2016generative}, and properties of partially-segmented point clouds~\cite{nash2017shape}, learning semantically meaningful data representations. On the other hand, GANs have been studied in the context of point set generation from images~\cite{fan2017point}, multi-view reconstruction~\cite{lin2018learning}, volumetric model synthesis~\cite{wu2016learning}, and, more recently, point cloud processing~\cite{achlioptas2018learning,li2018point}. However, neither of the mentioned approaches provides accurate multi-scale modelling of 3D representation with high resolution, which has been demonstrated to drastically improve image synthesis with GANs~\cite{denton2015deep}. Additionally, in several instances, for point cloud generation, some form of input (\eg, an image) is required~\cite{fan2017point}. In contrast, we are able to generate highly detailed point clouds unconditionally.

\noindent\textbf{Multi-scale neural networks on 3D shapes. }
\label{review:multiscale_lap_processing}
For 2D images, realistic synthesis with GANs has first been proposed in~\cite{denton2015deep} with a coarse-to-fine Laplacian pyramid image representation. Surprisingly, little work of similar nature has been done in multi-scale point cloud modelling. \cite{mandikal2019dense} propose a deep pyramid neural network for point cloud upsampling. However, their model accepts RGB images as input and is unable to synthesise novel 3D shapes. We, however, operate directly on unstructured 3D point sets and provide a full generative model for 3D shapes.  \cite{yifan2019patch} operate on patches in a progressive manner, thus implicitly representing a multi-scale approach. However, their model specifically tackles the point set upsampling task and cannot produce novel point clouds. In constrast, our generative model purposefully learns mutually coherent representations of full 3D shapes at multiple scales through the use of a latent-space Laplacian pyramid.

\section{Framework}
\label{sec:methods}

Our model for learning with 3D point clouds is inspired by works on Latent GANs~\cite{achlioptas2018learning} and Laplacian GANs~\cite{denton2015deep}. Therefore, we first briefly describe these approaches.

\begin{figure*}[t]
  \centering
  \includegraphics[width=\textwidth]
    {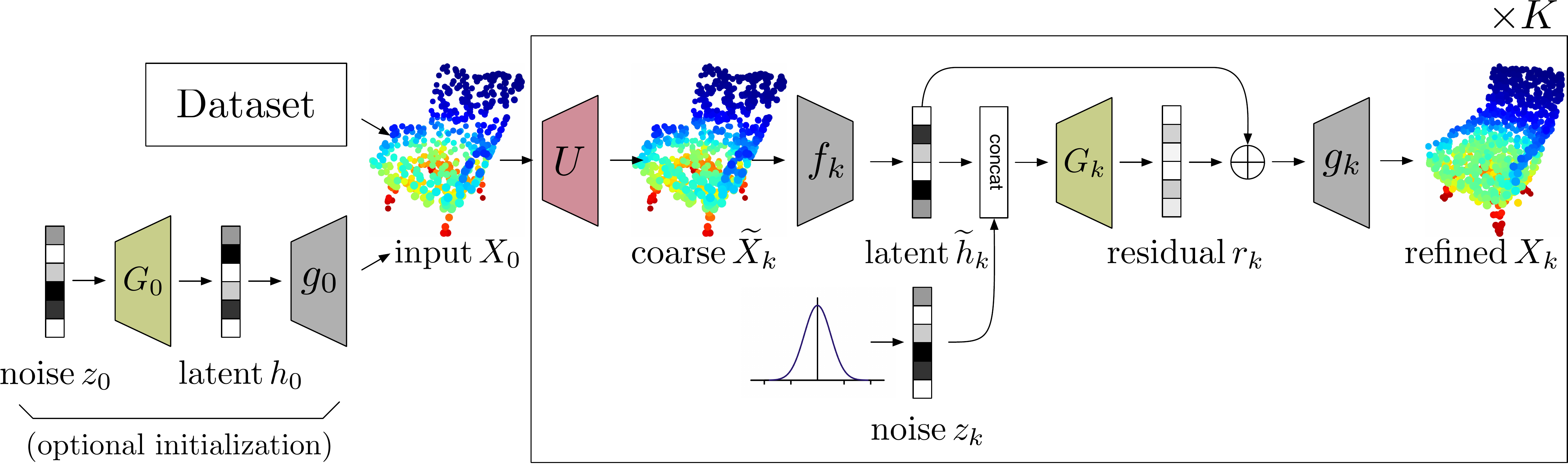}
  \caption{Full architecture of LSLP-GAN model. The network either accepts or generates an initial point cloud $X_0$ and processes it with a series of $K$ learnable steps. Each step (1) upsamples its input using a non-learnable operator $U$, (2) encodes the upsampled version into the latent space by $f_k$, (3) performs correction of the latent code via a conditional GAN $G_k$, and (4) decodes the corrected latent code using $g_k$.}
  \label{fig:framework-diagram}
\end{figure*}

\subsection{Latent GANs and Laplacian GANs}
\label{subsec:gans-review}

The well-known autoencoding neural networks, or autoencoders, compute \emph{latent codes} $h \in \mathbb{R}^d$ for an input object~$X \in \mathbb{X}$ through the means of an encoding network $f(X)$. This latent representation can later be decoded in order to reconstruct the original input, via a decoding network $g(h)$. \emph{Latent GANs}~\cite{achlioptas2018learning} are built around the idea that real latent codes commonly occupy only a subspace of their enclosing space $\mathbb{R}^d$, \ie live on a manifold embedded in $\mathbb{R}^d$. Thus, it should be possible to synthesise samples by learning the manifold of real latent codes, presumably an easier task compared to learning with the original high-dimensional representations. A GAN model is thus used in addition to an autoencoder model, yielding artificial latent codes obtained as a result of an adversarial game. Such models were recently applied to learning with point clouds, where they compare favourably with GANs that operate on the raw input~\cite{achlioptas2018learning}.

\emph{Laplacian GANs}~\cite{denton2015deep} increase image resolution in a coarse-to-fine manner during synthesis, aiming to produce high quality samples of natural images. This is motivated by the fact that standard GANs give acceptable results for generation of low-resolution images but fail for images of higher resolution. Laplacian GANs overcome this problem by a cascading image synthesis with a series of generative networks $G_0, \ldots, G_n$, where each network $G_k$ learns to generate a high-frequency residual image $r_k = G_k(U(I_{k+1}), z_k)$ conditioned on the upsampled image $I_k$ provided by $G_{k-1}$. Thus, an image at stage $k$ is represented via:
\begin{equation}
\label{eq:image_lappyramid}
    I_k = U(I_{k+1}) + G_k(U(I_{k+1}), z_k),
\end{equation}
where $U(\cdot)$ is an upsampling operator and $z_k$ is a noise vector.
For point clouds, a change in image resolution transforms to resampling, where subsets of points may be selected to form a low-resolution 3D shape, or reconstructed for a higher resolution 3D shape.

\subsection{Representation learning with latent-space Laplacian pyramids}
\label{subsec:pcu-gan-in-nutshell}

\noindent\textbf{Spaces of 3D point clouds. }
We start with a series of 3D shape spaces $\mathbb{R}^{n_0 \times 3}, \mathbb{R}^{n_1 \times 3}, \ldots, \mathbb{R}^{n_K \times 3}$ with $n_0 < n_1 < \ldots < n_K$ (specifically, we set $n_k = 2^k \cdot n_0$). A 3D shape can be represented in the space $\mathbb{R}^{n_k \times 3}$ with a 3D point sampling of its surface $X_k = \{x_i\}_{i=1}^{n_k}$. If this sampling maintains sufficient uniformity for all $k$, then the sequence of 3D point clouds $X_0, X_1, \ldots$ represents a progressively finer model of the  surface. Our intuition in considering the series of shape spaces is that modelling highly detailed 3D point clouds represents a challenge due to their high dimensionality. Thus, it might be beneficial to start with a low-detail (but easily constructed) model $X_0$ and decompose the modelling task into a sequense of more manageable stages, each aimed at a gradual increase of detail.



\noindent\textbf{Training auto-encoding point networks on multiple scales. }
Learning the manifold of latent codes has been demonstrated to be beneficial in terms of reconstruction quality~\cite{achlioptas2018learning}. Motivated by this observation, we use 3D shape spaces $\{ \mathbb{R}^{n_k \times 3} \}_{k=1}^K$ and construct a series of corresponding latent spaces $\{ \mathbb{R}^{d_k} \}_{k=1}^K$ by training point autoencoders $\{(f_k, g_k)\}_{k = 1}^K$. Note that an autoencoder $(f_k, g_k)$ is trained using the resolution $n_k$ of 3D point clouds, which grows as $k$ increases. As our method strongly relies on the quality of autoencoders, we evaluate reliability of their mappings from 3D space into latent space in Section~\ref{sec:experiments}. After training the autoencoders, we fix their parameters and extract latent codes for shapes in each of the 3D shape spaces. 

\begin{figure}[!h]
  \centerline{
  \includegraphics[width=\columnwidth]
    {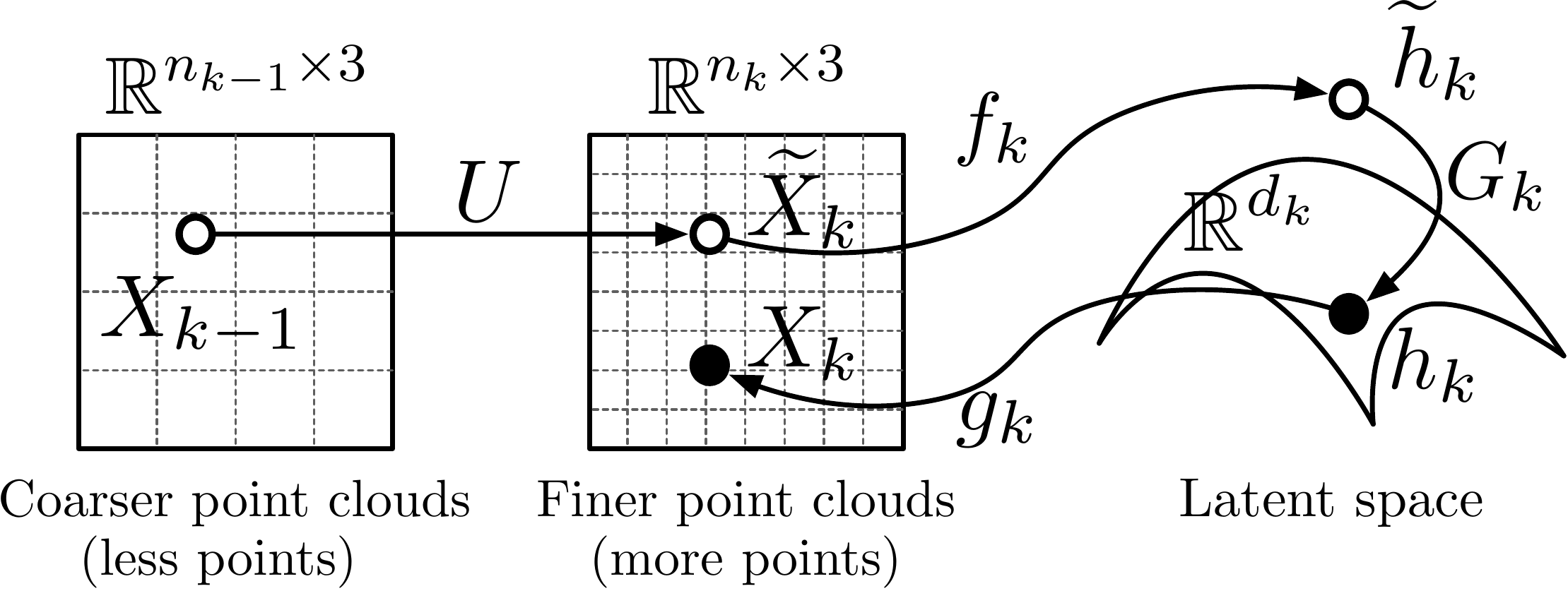}
  }
  \caption{A detailed operation scheme of our latent-space Laplacian pyramid (see the accompanying text).}
  \label{fig:latent-space-mechanics}
\end{figure}

\noindent\textbf{Laplacian pyramid in the spaces of latent codes. }
For what follows, it is convenient to assume that we are given as input a point cloud $X_{k-1} \in \mathbb{R}^{n_{k-1} \times 3}$. 
We aim to go from $X_{k-1}$ to $X_k$, \ie to increase resolution from $n_{k-1}$ to $n_k = 2 n_{k-1}$, generating additional point samples on the surface of an underlying shape. Figure~\ref{fig:latent-space-mechanics} illustrates our reasoning schematically. 

We start by processing the input point cloud by a simple upsampling operator $U(\cdot)$, obtaining a coarse point cloud $\widetilde{X}_k = U(X_{k-1})$: for each point $x \in X_{k-1}$ we create a new instance $\widetilde{x} = \frac {1} {m} \sum_{i \in \text{NN}(x)} x_i$ where $\text{NN}(x)$ is a set of $m$ nearest Euclidean neighbours of $x$ in $X_{k-1}$ (we use $m = 7$ neighbours, including $x$), and add it to the point cloud. This procedure represents a simple linear interpolation and forms exactly $n_k$ points located in the vicinity of the real surface. However, the computed point cloud $\widetilde{X}_k$ generally contains perturbed points, and we view it only as a rough approximation to our desired $X_k$.

We map the coarse point cloud $\widetilde{X}_k$ by $f_k$ into a latent code $\widetilde{h}_k = f_k(\widetilde{X}_k)$, which we assume to be offset by a small delta from the manifold of latent representations due to an interpolation error in $\widetilde{X}_k$.
To compensate for this offset in the latent space, we compute an additive correction $r_k$ to~$\widetilde{h}_k$ using a generator network $G_k$, resulting in a corrected code $h_k = \widetilde{h}_k + r_k = \widetilde{h}_k + G_k(\widetilde{h}_k, z_k)$. 
Decoding $h_k$ by $g_k$, we obtain a refined point cloud $X_k = g_k(h_k)$ with resolution $n_k$.

Putting together the full procedure in the space of latent representations leads to a series of relations
\begin{equation}
\label{fig:latent_lappyramid}
    h_k = f_k(U(X_{k-1})) + G_k(f_k(U(X_{k-1})), z_k),
\end{equation}
which is a latent-space equivalent of~\eqref{eq:image_lappyramid}. Hence, we call the resulting series $\{h_k\}_{k=0}^K$ of  hierarchical representations a \emph{latent-space Laplacian pyramid} (LSLP).


\noindent\textbf{Training latent GANs on multiple scales. } 
To perform meaningful corrections of the rough latent code $\widetilde{h}_k$, each generator $G_k$ faces a~challenge of learning the subtle differences between the latent codes of true high-resolution 3D point clouds and those of coarsely upsampled ones. Thus, we train a series of latent GANs $\{(G_k, D_k)\}_{k = 1}^K$ by forcing the generator $G_k$ to synthesise residuals $r_k$ in the latent space conditioned on the input $\widetilde{h}_k$, and the discriminator $D_k$ to distinguish between the real latent codes $h_k \in \mathbb{R}^{d_k}$ and the synthetic ones $\widetilde{h}_k + G_k(\widetilde{h}_k, z_k)$. Note that as each (but the first) latent GAN accepts a rough latent code $\widetilde{h}_k$, they may be viewed as conditional GANs (CGANs)~\cite{mirza2014conditional}.

\noindent\textbf{Two execution modes: synthesis and upsampling. }
In the text above, we assumed an initialiser $X_0 \in \mathbb{R}^{n_0 \times 3}$ to be given as an input, which is the case in particular applications, such as upsampling or shape completion. However, our framework can as easily function in a purely generative mode, sampling unseen high-resolution point clouds on the fly. To enable this, we start with an (unconditional) latent GAN $G_0$ and produce a point cloud $X_0 = g_0(G_0(z_0))$, which serves as an input to the remaining procedure.



An overview of our architecture is presented in~Figure \ref{fig:framework-diagram}.


\noindent\textbf{Architectural and training details of our framework. }
The architecture of all our networks is based on the one proposed in~\cite{achlioptas2018learning}, where the autoencoders follow the PointNet~\cite{Pointnet} encoders design and have fully-connected decoders, and GANs are implemented by the MLPs. 



When training the autoencoders, we optimise the Earth Mover's Distance (EMD) given by:
\[
d_{\text{EMD}}(X, Y) = \min\limits_{\phi: X \to Y} \sum\limits_{x \in X} || x - \phi(x) ||_2
\]
where $\phi$ is a bijection, obtained as a solution to the optimal transportation problem involving the two point sets $X, Y \in \mathbb{R}^{n_k \times 3}$. Training the GANs is performed by optimising the commonly used objectives~\cite{goodfellow2014generative,mirza2014conditional}.

\section{Evaluation and applications}
\label{sec:experiments}

\subsection{Setup of our evaluation}
\label{exper:setup_of_evaluation}


\noindent\textbf{Training datasets. }
For all our experiments, we use the meshes from the ShapeNet dataset~\cite{chang2015shapenet}. We have perfomed experiments using separate \emph{airplane, table, chair, car}, and \emph{sofa} classes in the Shapenet dataset, as well as using a \textit{multi-class} setup.
We train three stages of autoencoders and generative models on resolutions of $n_0 = 512$, $n_1 = 1024$, and $n_2 = 2048$ points, respectively, using a training set of 3046, 7509, 5778, 6497, and 4348 3D shapes for the classes \textit{airplane, table, chair, car,} and \textit{sofa,} respectively, and 9000 shapes for our \textit{multi-class} setup. 
We have used Adam~\cite{kingma2014adam} optimisers to train both autoencoders and GANs. All autoencoders have been trained for 500~epochs with the initial learning rates of $5 \times 10^{-4}$, $\beta_1 = 0.9$ and a~batch size of 50; GANs have been trained for 200~epochs with the initial learning rates of $10^{-4}$, $\beta_1 = 0.9$ and a~batch size of 50.

\noindent\textbf{Metrics. }
Along with the Earth Mover's Distance (EMD), we assess the point cloud reconstruction performance using the Chamfer Distance (CD) as a second commonly adopted measure, given by
\[
d_{\text{CD}}(X, Y) = \sum\limits_{x \in X} \min\limits_{y \in Y} || x - y ||_2^2 + \sum\limits_{y \in Y} \min\limits_{x \in X} || x - y ||_2^2.
\]
To evaluate the generative models, we employ the Jensen-Shannon Divergence (JSD), coverage (COV), and Minimum Matching Distance (MMD) measures, following the scheme proposed in~\cite{achlioptas2018learning}. JSD is defined over two empirical measures $P$ and $Q$ as $\JSD{P}{Q} = 1/2 \KL{P}{M} + 1/2 \KL{Q}{M}$ where $\KL{P}{Q}$ is the Kullback-Leibler divergence between the measures $P$ and $Q$, $M = 1/2 (P + Q)$, and measures $P$ and $Q$ count the number of points lying within each voxel in a voxel grid across all point clouds in the two sets $A$ and $B$, respectively. COV measures a fraction of point clouds in $B$ approximately represented in $A$; to compute it, we find the nearest neighbours in $B$ for each $x \in A$. MMD reflects the fidelity of the set $A$ with respect to the set $B$, matching every point cloud of $B$ to the one in $A$ with the minimum distance and computing the average of distances in the matching. 

\subsection{Experimental results}

\noindent\textbf{Evaluating autoencoders. }
We first evaluate our point autoencoders to validate their ability to compute efficient latent representations with increasing resolutions of input 3D point clouds. To compute the reconstructions, we encode into the latent space and decode back the 3D shapes from the test split of the respective class unseen during training. In Table~\ref{table:autoencoder_metrics} we display the reconstruction quality of our auto-encoders for the three levels of resolution, using CD and EMD measures. As the sampling density increases, both measures improve as expected.

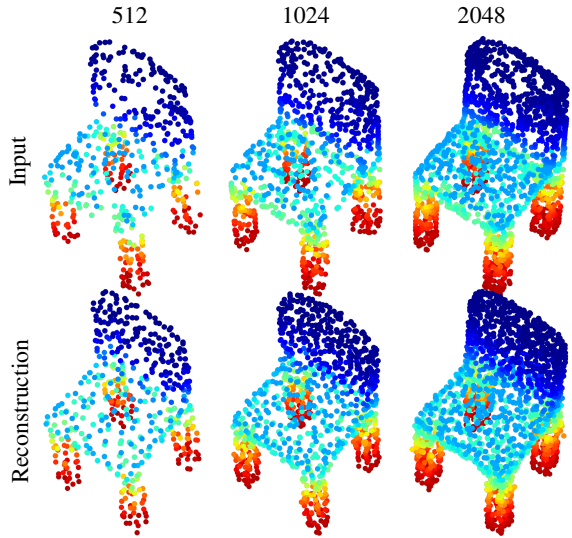
\begin{figure}[h!]
  \pgfmathsetmacro{\imgwidth}{\linewidth-\figspace}%
  \centerline{\small\begin{tikzpicture}
                \node[inner sep=0] (image) {\includegraphics[width=\imgwidth pt]{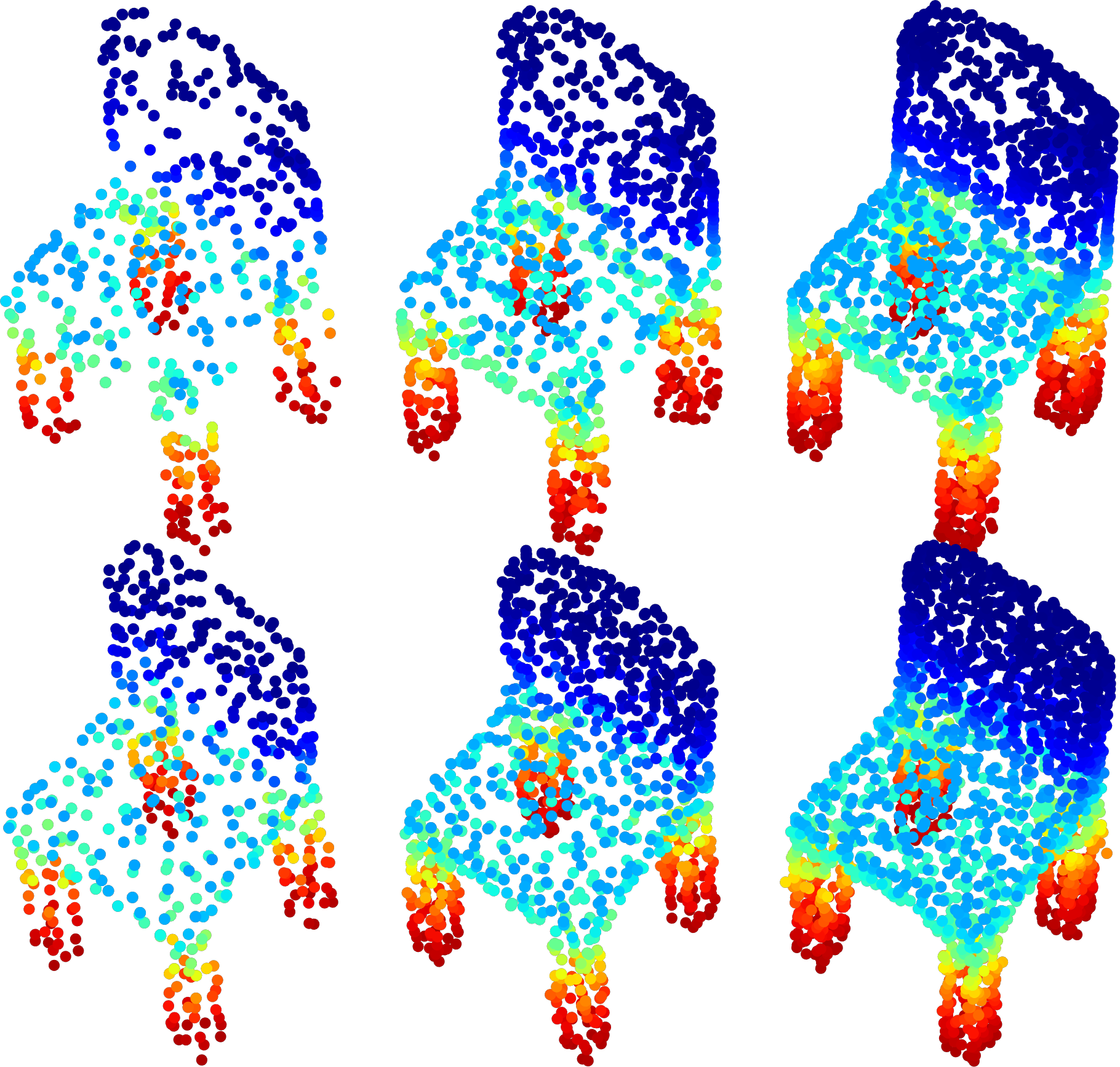}};
                \begin{scope}[shift=(image.north west), x={(image.north east)},y={(image.south west)}]
                  \node[rotate=90] at (-\figspace/2, .25) {Input};
                  \node[rotate=90] at (-\figspace/2, .75) {Reconstruction};
                  \node at (.167, \figspace/2) {512};
                  \node at (.5, \figspace/2) {1024};
                  \node at (.833, \figspace/2) {2048};
                \end{scope}
  \end{tikzpicture}}
  \caption{Inputs and reconstructions using our autoencoders at resolutions
$n_i \in \{512, 1024, 2048\}$ of the 3D point cloud.}
  \label{fig:autoencoder_gt_dec}
\end{figure}

\begin{table}[h]
\centering
\resizebox{\columnwidth}{!}{%
\begin{tabular}{@{}lllllll@{}}
\toprule
\multirow{2}{*}{\begin{tabular}[c]{@{}l@{}}Shape\\ class\end{tabular}} & \multicolumn{3}{c}{CD $\times 10^{-3}$} & \multicolumn{3}{c}{EMD $\times 10^{-3}$} \\ 
 & \multicolumn{1}{c}{512} & \multicolumn{1}{c}{1024} & \multicolumn{1}{c}{2048} & \multicolumn{1}{c}{512} & \multicolumn{1}{c}{1024} & \multicolumn{1}{c}{2048} \\ \midrule
\emph{chair}     & 0.16 & 0.10 & 0.07 & 60.2 & 53.5 & 48.3 \\
\emph{airplane}  & 0.57 & 0.38 & 0.29 & 39.4 & 34.5 & 30.8 \\
\emph{table}     & 1.41 & 0.96 & 0.67 & 56.9 & 50.1 & 45.6 \\ \bottomrule
\end{tabular}%
}
\caption{Reconstruction quality using our autoencoders at resolutions
$n_i \in \{512, 1024, 2048\}$ of the 3D point cloud. }
\label{table:autoencoder_metrics}
\end{table}

Figure~\ref{fig:autoencoder_gt_dec} demonstrates the ground-truth and decoded 3D point clouds, respectively, at all stages in our autoencoders. We conclude that our models can represent the 3D point clouds at multiple resolutions.

\noindent\textbf{Evaluating generative models. }
We further evaluate our generative models using the MMD-CD, COV, and JSD measures, in both single-class and multi-class setups. To this end, we train our LSLP-GAN using the latent spaces obtained with the previously trained autoencoders. Table~\ref{table:gan_metrics_single_class} compares our LSLP-GAN and the L-GAN model~\cite{achlioptas2018learning}. We consistently outperform the baseline L-GAN across all object classes according to the quality metrics defined aboveß.


\begin{table}[h]
\centering
\resizebox{\columnwidth}{!}{%
\begin{tabular}{@{}lrrrrrr@{}}
\toprule
\multirow{2}{*}{\begin{tabular}[c]{@{}l@{}}Shape\\ class\end{tabular}} & \multicolumn{2}{c}{MMD-CD $\times 10^{-3}$}                              & \multicolumn{2}{c}{COV-CD, \%}                                 & \multicolumn{2}{c}{JSD $\times 10^{-3}$}                                 \\
& \multicolumn{1}{c}{L-GAN} & \multicolumn{1}{c}{Ours} & \multicolumn{1}{c}{L-GAN} & \multicolumn{1}{c}{Ours} & \multicolumn{1}{c}{L-GAN} & \multicolumn{1}{c}{Ours} \\ \midrule
\emph{car}       &  0.81  &  \textbf{0.71}  &  23.5  &  \textbf{32.1}  &  28.9  &  \textbf{24.2}\\
\emph{chair}      & 1.79  &  \textbf{1.71}  &  44.9  &  \textbf{47.8}  &  13.0  &  \textbf{10.1}\\
\emph{sofa}      &  1.26  &  \textbf{1.23}  &  43.9  &  \textbf{46.3}  &  9.6  &  \textbf{9.3}\\
\emph{table}     &  1.93  &  \textbf{1.77}  &  39.7  &  \textbf{47.8}  &  19.9  &  \textbf{10.1}\\
\emph{airplane}  &  0.53  &  \textbf{0.51}  &  41.7  &  \textbf{44.0}  &  17.1  &  \textbf{13.8}\\
\emph{multiclass} & 1.66  &  \textbf{1.55}  &  41.4  &  \textbf{45.7}  &  14.3  &  \textbf{9.8}\\ \bottomrule
\end{tabular}
}
\caption{Performance evaluation of our proposed LSLP-GAN model as compared to the baseline L-GAN model~\cite{achlioptas2018learning}.}
\label{table:gan_metrics_single_class}
\end{table}

\begin{figure*}[tp]
    \pgfmathsetmacro{\imgwidth}{\linewidth-\figspace}%
    \centerline{\small\begin{tikzpicture}
                \node[inner sep=0] (image) {\includegraphics[width=\imgwidth pt]{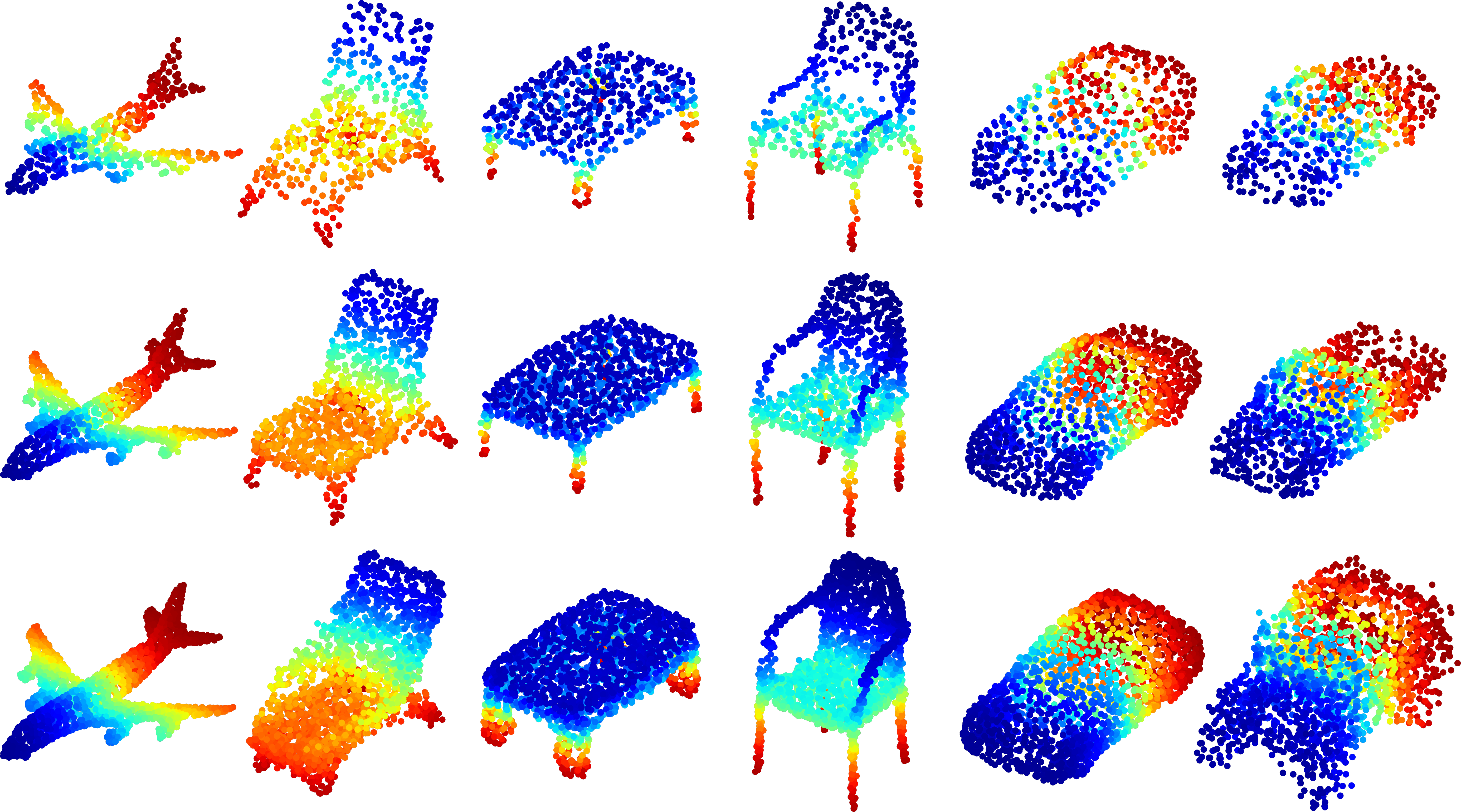}};
                \begin{scope}[shift=(image.north west), x={(image.north east)},y={(image.south west)}]
                    \node[rotate=90] at (-\figspace/2, .167) {512};
                    \node[rotate=90] at (-\figspace/2, .5) {1024};
                    \node[rotate=90] at (-\figspace/2, .833) {2048};
                    \draw (.5,0) -- (0.5,1);
                    \node at (0.25, \figspace/2) {Single class};
                    \node at (0.75, \figspace/2) {Multi-class};
                \end{scope}
    \end{tikzpicture}}
  \caption{Examples of shapes synthesised using our LSLP-GAN model. \textit{Left:} airplanes, chairs, and tables synthesised using our single-class models. \textit{Right:} samples of 3D shapes synthesised using our multi-class model, note that the overall geometry of the shape changes slightly due to averaging over many classes. The rightmost figure displays a failure mode for our model.}
  \label{fig:gan_novel_shape_synthesis}
\end{figure*}
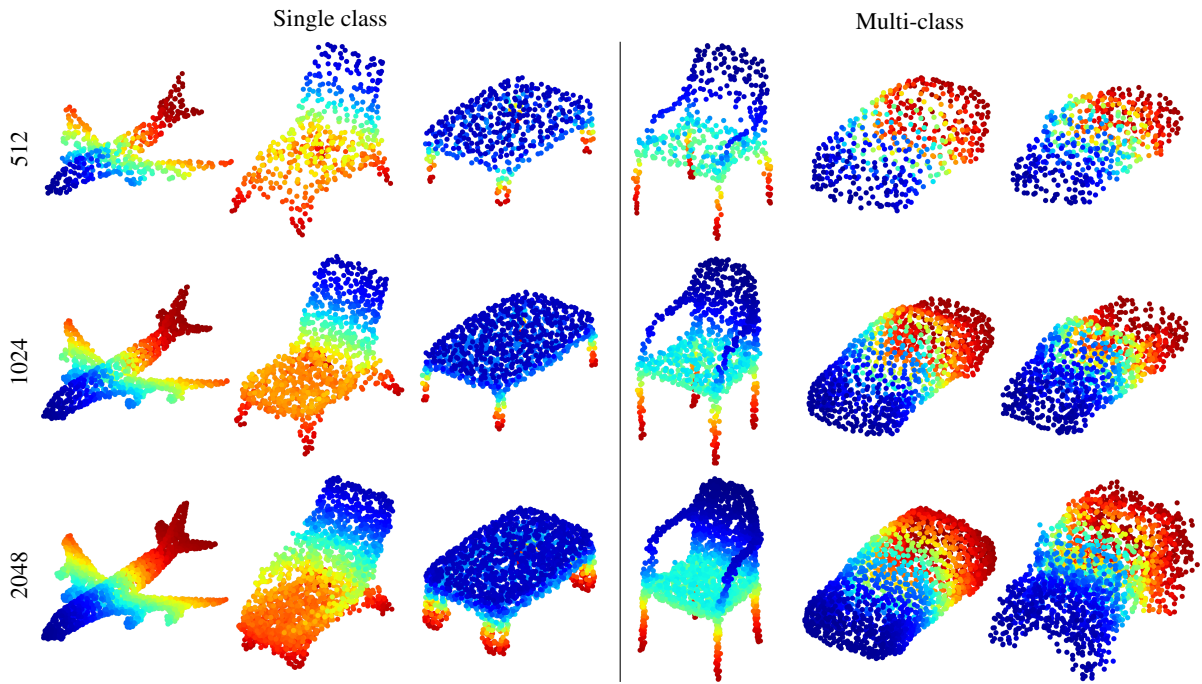

To demonstrate examples of novel 3D shape syntheses using our framework, we sample $z_0$ and process it with our framework, obtaining 3D point clouds $X_0, X_1, X_2$, which we display in Figure~\ref{fig:gan_novel_shape_synthesis}. Our framework can synthesise increasingly detailed 3D shapes, gradually adding resolution using a series of generative models.

\begin{figure}[ht]
  \centerline{\small\begin{tikzpicture}
                        \node[inner sep=0] (image) {\includegraphics[width=\linewidth]{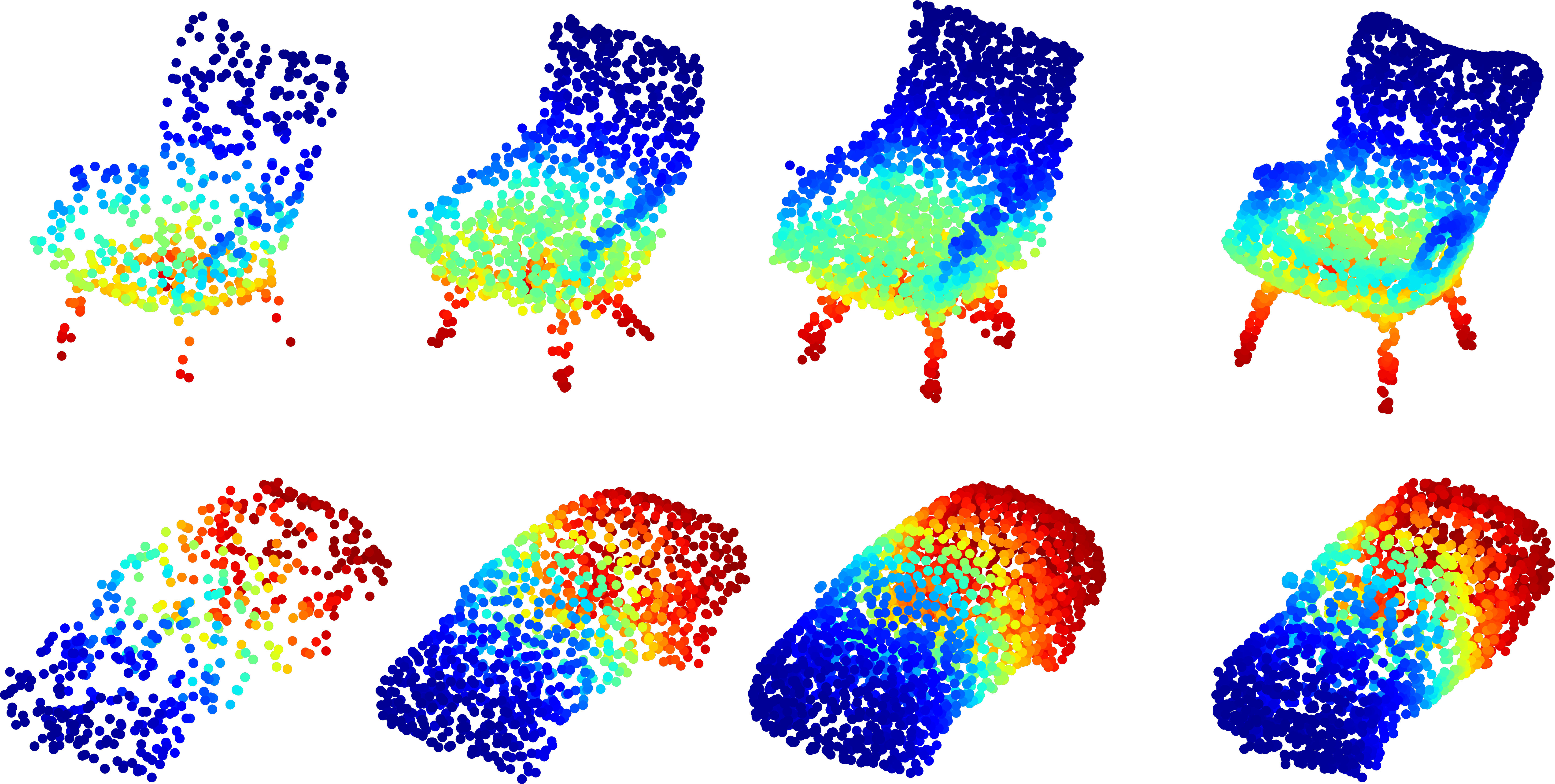}};
                        \begin{scope}[shift=(image.north west), x={(image.north east)},y={(image.south west)}]
                            \node at (.125, \figspace/2) {Input, 512};
                            \node at (.375, \figspace/2) {1024};
                            \node at (.625, \figspace/2) {2048};
                            \node at (.875, \figspace/2) {Reference};
                            \draw (.75,0) -- (0.75,1);
                        \end{scope}
  \end{tikzpicture}}
  \caption{3D point clouds upsampling results using our model, initialised with the input shape.}
  \label{fig:upsampling_application}
\end{figure}
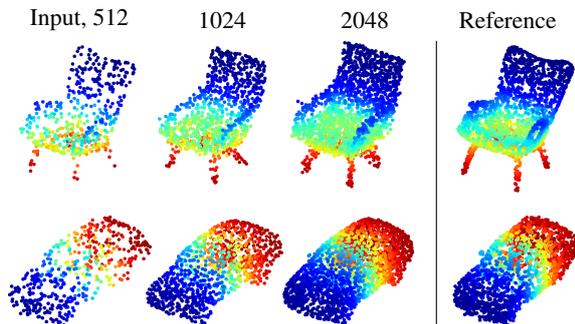

\noindent\textbf{Point cloud upsampling.}
\label{subsec:experiments-upsampling}
Generative models such as L-GAN and our proposed LSLP-GAN are a natural fit for the task of 3D point set upsampling, as they learn to generate novel points given the lower resolution inputs. Thus, we evaluate our framework by modelling the upsampling task using the low-resolution 3D shapes from the Shapenet dataset. We supply LSLP-GAN with a low-resolution point cloud from the test split of the multi-class dataset and increase its resolution four-fold from $n_0 = 512$ to $n_2 = 2048$ points, performing conditional generation using $G_1$ and $G_2$. Figure~\ref{fig:upsampling_application} displays 3D shapes upsampled using our multi-class model. Note that model has not been trained to perform upsampling directly, \ie to preserve the overall shape geometry when producing novel points, hence the subtle changes in 3D shapes as the upsampling progresses.




\section{Conclusion and future work}
\label{sec:conclusion}

We have presented LSLP-GAN, a novel deep adversarial representation learning framework for 3D point clouds. The initial experimental evaluation reveals the promising properties of our proposed model. We plan to \begin{inparaenum}[(i)] \item further extend our work, considering deeper pyramid levels and larger upsampling factors (\eg $\times 64$), and \item evaluate our framework using more challenging tasks such as shape completion. \end{inparaenum}

 \section*{Acknowledgement}

 The work of Youyi Zheng is partially supported by the National Key Research \& Development Program of China (2018YFE0100900). The work of Vage Egiazarian, Alexey Artemov, Oleg Voynov and Evgeny Burnaev is supported by The Ministry of Education and Science of Russian Federation, grant No. 14.615.21.0004, grant code: RFMEFI61518X0004. The work of Luiz Velho is supported by CNPq/MCTIC/BRICS-STI No 29/2017 --- Grant No: 442032/2017-0. The authors Vage Egiazarian, Alexey Artemov, Oleg Voynov and Evgeny Burnaev acknowledge the usage of the Skoltech CDISE HPC cluster Zhores for obtaining results presented in this paper.

\FloatBarrier

\bibliographystyle{apalike}
\bibliography{bib/bibliography.bib}

\end{document}